\documentclass[conference]{IEEEtran}
\usepackage{cite}
\usepackage{amsmath,amssymb,amsfonts}
\usepackage{algorithmic}
\usepackage{graphicx}
\usepackage{textcomp}
\usepackage{xcolor}
\usepackage{multirow}
\usepackage{booktabs}
\usepackage{array}
\usepackage{makecell}
\usepackage{url}
\usepackage[hidelinks]{hyperref}

\def\BibTeX{{\rm B\kern-.05em{\sc i\kern-.025em b}\kern-.08em
    T\kern-.1667em\lower.7ex\hbox{E}\kern-.125emX}}
\begin{document}

\bstctlcite{BSTcontrol}
\setlength{\textfloatsep}{10pt plus 1.0pt minus 2.0pt}
\setlength{\floatsep}{8pt plus 1.0pt minus 2.0pt}
\setlength{\intextsep}{10pt plus 1.0pt minus 2.0pt}
\title{Aligning LLMs for the Classroom with Knowledge- Based Retrieval: A Comparative RAG Study\\}

\author{
\IEEEauthorblockN{Amay Jain}
\IEEEauthorblockA{\textit{Student} \\
\textit{Downingtown STEM Academy}\\
Downingtown, PA \\
jain.amay04@gmail.com}
\and
\IEEEauthorblockN{Liu Cui}
\IEEEauthorblockA{\textit{Department of Computer Science} \\
\textit{West Chester University of Pennsylvania}\\
West Chester, PA \\
lcui@wcupa.edu}
\and
\IEEEauthorblockN{Si Chen}
\IEEEauthorblockA{\textit{Department of Computer Science} \\
\textit{West Chester University of Pennsylvania}\\
West Chester, PA \\
schen@wcupa.edu}
}

\maketitle

\begingroup
\renewcommand\thefootnote{}\footnotetext{\textit{Preprint notice—} This work has been submitted to the IEEE for possible publication. Copyright may be transferred without notice, after which this version may no longer be accessible.}
\addtocounter{footnote}{-1}
\endgroup

\begin{abstract}

Large language models like ChatGPT are increasingly used in classrooms, but they often provide outdated or fabricated information that can mislead students. Retrieval Augmented Generation (RAG) improves reliability of LLMs by grounding responses in external resources. We investigate two accessible RAG paradigms, vector-based retrieval and graph-based retrieval to identify best practices for classroom question answering (QA). Existing comparative studies fail to account for pedagogical factors such as educational disciplines, question types, and practical deployment costs. Using a novel dataset, EduScopeQA, of 3,176 questions across academic subjects, we measure performance on various educational query types, from specific facts to broad thematic discussions. We also evaluate system alignment with a dataset of systematically altered textbooks that contradict the LLM’s latent knowledge. We find that OpenAI Vector Search RAG (representing vector-based RAG) performs well as a low-cost generalist, especially for quick fact retrieval. On the other hand, GraphRAG Global excels at providing pedagogically rich answers to thematic queries, and GraphRAG Local achieves the highest accuracy with the dense, altered textbooks when corpus integrity is critical. Accounting for the 10-20x higher resource usage of GraphRAG (representing graph-based RAG), we show that a dynamic branching framework that routes queries to the optimal retrieval method boosts fidelity and efficiency. These insights provide actionable guidelines for educators and system designers to integrate RAG-augmented LLMs into learning environments effectively.

\end{abstract}

\begin{IEEEkeywords}
Large language models, Educational technology, Retrieval augmented generation
\end{IEEEkeywords}

\section{Introduction}
Large‐language models (LLMs) have rapidly entered secondary and post‑secondary classrooms, promising adaptive tutoring and richer instructional material. However, educators consistently report a critical obstacle that limits dependable classroom use: \textbf{misalignment with educational goals}. \cite{education_ai}. 

LLMs struggle with hallucinated or fabricated content and responses misaligned with curriculum standards. Their internet-scale training can introduce tangential details that, while sometimes helpful, may confuse students or undermine course coherence \cite{hallucinations_ai}. \textbf{Knowledge shifts} further complicate this: school curricula are periodically updated, and facts or conventions can change (historical interpretations, scientific nomenclature, etc.). An LLM trained on older information can supply superseded facts or methodologies.\cite{gpt_education}

Retrieval-Augmented Generation (RAG)\cite{rag_lewis}, an approach that supplements generative models with knowledge retrieval, has emerged as a promising strategy to address these issues\cite{finance_rag}. Two specific RAG variants, \textbf{vector-based RAG} and \textbf{graph-based RAG}, have gained popularity.

Vector-based RAG is the prevailing standard for retrieval augmentation, using embedded vector similarity\cite{rag_survey}. Graph-based RAG extends this by incorporating graph-structured knowledge bases, enabling nuanced multi-hop retrieval and information synthesis\cite{microsoft}. While prior work has compared these methods,  their relative merits for \textit{curriculum-aligned classroom question anwering} remain under-explored.

Recently, other RAG methods have emerged (sparse, hybrid, neural symbolic, etc.), but in this paper, we focus on vector-based and graph-based RAG, due to their minimal infrastructure requirements and ability to leverage off-the-shelf APIs. For classroom adoption, ease of use is paramount. Thus, solutions requiring minimal setup, maintenance are preferred. 

To ground our comparison in practical tools, we focus on two widely popular turnkey RAG solutions that are already accessible to most educators: OpenAI Vector File Search (``OpenAI RAG")\cite{openai_retrieval} as our vector-based representative, and Microsoft’s GraphRAG framework, available in both Local and Global modes, as our graph-based representative.

Since the needs of educators and students differ across disciplines, scopes, and tasks, we design two complementary evaluations. In \textbf{Case Study 1} (CS1), we test each system's ability to retrieve and synthesize information on 3,176 QA pairs at different levels of granularity across four academic subjects. In \textbf{Case Study 2} (CS2), we leverage the KnowShiftQA dataset\cite{knowshift}, which contains 3,005 QA pairs from textbooks in which key facts have been systematically altered, to assess whether each retrieval method can accurately provide updated information (from the texts) rather than falling back on an LLM's outdated or common knowledge. 

Beyond accuracy, we measure indexing cost, response latency, and scalability, factoring the typical constraints of schools and universities. 

Our evaluation addresses \textbf{three key research questions} for educational deployment:

\vspace{0.3em}
\noindent\textbf{Research Questions.}
\begin{enumerate}
  \item In multi‑disciplinary classroom QA, how do vector-based RAG and graph-based RAG differ with respect to answer accuracy and explanation quality?
  \item When curricular knowledge shifts, which method better resists outdated or superseded information?
  \item Under typical school constraints, how do these methods compare in resource efficiency?
\end{enumerate}

\vspace{0.3em}
\noindent This work offers three main \textbf{contributions}:
\begin{enumerate}
\item a novel, multi-subject, multi-scope QA dataset focused on secondary and post-secondary educational applications---EduScopeQA;
\item a comparative evaluation of vector-based vs. graph-based RAG methods against educational factors like subject, question type, text size, and resource constraints;
\item a proof of concept showing how each system's strengths found in the evaluation can guide practical deployment and usage of LLM-based QA systems in education.
\end{enumerate}

\section{Related Work}

\textbf{Vector-based RAG} \cite{rag_survey} chunks and vector-encodes a provided corpus using a neural embedding model. The user query is similarly embedded, and the most semantically similar chunks from the corpus are retrieved and passed to the LLM with the query to produce grounded answers\cite{rag_lewis}. There is a growing body of research exploring vector-based RAG in educational tasks. \cite{math_rag} found that vector-based RAG improved the traceability of mathematical solutions, and \cite{rag_grader} used vector-based RAG to boost automated grading accuracy.

OpenAI’s Vector File Search API (OpenAI RAG) uses a managed vector store to perform semantic search over uploaded files. It automates chunking, embedding, and retrieval within a single API call. With these abstractions, OpenAI’s pipeline dramatically simplifies the use of vector-based RAG in everyday use cases such as education. Given OpenAI ChatGPT's mainstream adoption, especially in education, \cite{chatgpt_enchancelearning}, evaluating OpenAI RAG provides practical relevance.

\textbf{Graph-based RAG} uses an LLM to organize documents into a structured knowledge graph. The system identifies key people, places, and concepts, and constructs a graph with these identified entities as nodes and their relationships as edges. The LLM then produces summaries for clusters of entities. Thus, at query time, graph-based RAG traverses this graph, aggregating evidence from the relevant nodes\cite{microsoft}.   

There are different retrieval configurations for graph-based RAG that depend on the locality of the graph traversal. ``Local" community search retrieves node‑centred subgraphs and their immediate community summaries, whereas ``Global" search aggregates summaries across all communities, spanning the entire document’s knowledge structure. Local emphasizes precision, while global values coverage\cite{microsoft}.

Although many variants of graph-based RAG exist, such as LightRAG\cite{lightrag} and LazyRAG, we use Microsoft’s GraphRAG as a representative due to its open-source nature, ease of use, and comparable performance, which makes it widely accessible to educators.

Several studies have compared vector and graph-based RAG, but on non-educational corpora and with evaluations not focused on curriculum-aligned QA. In GraphRAG’s initial paper, global graph retrieval achieved comprehensiveness win rates of 72–83\% on podcast transcripts and 72–80\% on news articles, outpacing vector RAG, yet these tasks emphasize broad summarization metrics (e.g., ROUGE, diversity) rather than pinpointed, QA accuracy on textbook content\cite{microsoft}. 

However, GraphRAG’s strengths in large datasets remain unclear. One study \cite{rag_vs_graph} found that GraphRAG Local/Global actually underperforms compared to vector RAG on large narrative datasets such as NovelQA by 4\%-18\%, whereas \cite{neutral_evaluation} found that on the Ultra Domain Benchmark, GraphRAG outpaces vector RAG as dataset size increases.

One significant barrier to a pedagogical evaluation is that existing QA datasets such as the NQ dataset\cite{nq_dataset}, HotPotQA\cite{hotpotQA}, and MultiHopRAG\cite{multihop} draw on fairly homogeneous sources such as Wikipedia or news articles, not reflecting the varied styles, structures, and lengths of real educational texts. Curriculum-aligned datasets exist mainly for scientific knowledge. CK‑12 QA (TQA)\cite{ck-12QA} includes 12,000 middle‑school science multiple‑choice items linked to textbook excerpts. OpenBookQA\cite{openbookQA} contains 5,957 elementary‑level science multiple-choice questions paired with facts. However, these mostly focus on shorter texts, and do not span a broad range of educational disciplines or support explanatory questions—gaps which our new dataset aims to fill.

In light of these works, we build on prior evaluations by focusing on realistic educational texts, varying lengths, subjects, and question types. Additionally, we account for pedagogical value (directness and learnability), knowledge integrity under shifts, and resource constraints of real schools.

\section{Case Study 1 (CS1)}

\subsection{Novel Dataset - EduScopeQA}
We constructed a comprehensive dataset of 3,176 QA pairs and their respective texts spanning four academic subjects—History, Literature, Science, and Computer Science—for a combined size of 2.1 million tokens ($\approx$ 500,000 tokens per subject). We intentionally selected texts resembling those studied in schools or universities.

For \textbf{Literature} and \textbf{History}, we picked some of Project Gutenberg's\cite{gutenberg} most downloaded works in their categories: two lengthy fictional novels and a variety of primary and secondary historical texts. \textbf{Science} is a full-length textbook, chosen for its structured organization of facts, processes, and terminology, and the \textbf{Computer Science} set is composed of highly-cited technical monographs from arXiv\cite{arxiv}. Each subject's corpus differs from each other in lexical and syntactical complexity, fluidity, and density of information, mimicking real classrooms, and allowing for a complete assessment of the strengths and limitations of RAG systems.

More information and citations can be seen in Table~\ref{tab:dataset_composition}. The dataset and a download script for texts is available at:
\mbox{\url{https://github.com/Amay-J/EduScopeQA}}. This release covers History,
Literature, and Science. Licensing information appears in the README.

\begin{table}[htbp]
    \centering
    \caption{EduScopeQA Dataset Composition}
    \label{tab:dataset_composition}
    \resizebox{\columnwidth}{!}{
    \renewcommand{\arraystretch}{1.3}
    \begin{tabular}{|p{0.07\textwidth}|>{\raggedright\arraybackslash}p{0.2\textwidth}|r|r|r|r|}
    \hline
    \textbf{Subject} & \textbf{Source Text} & \textbf{Words} & \textbf{Specific} & \textbf{Sectional} & \textbf{Thematic} \\
    \hline
    \multirow{3}{0.07\textwidth}{\textbf{Literature}   \cite{melville_mobydick}\cite{alcott_littlewomen}}
    & Moby-Dick; or, The Whale & 208,465 & 356 & 40 & 20 \\
    & Little Women & 188,683 & 323 & 40 & 20 \\
    & \textit{Subtotal} & \textit{397,148} & \textit{679} & \textit{80} & \textit{40} \\
    \hline
    \multirow{7}{0.07\textwidth}{\textbf{History} \cite{peary_northpole}\cite{barrows_philippines}\cite{franklin_autobiography}\cite{keynes_economic_consequences}\cite{douglass_narrative}\cite{paine_common_sense}}
    & The North Pole: Its Discovery & 97,980 & 173 & 20 & 10 \\
    & A History of the Philippines & 77,475 & 173 & 16 & 8 \\
    & Autobiography of Benjamin Franklin & 76,108 & 134 & 15 & 7 \\
    & Economic Consequences of the Peace & 69,966 & 123 & 14 & 7 \\
    & Life of Frederick Douglass & 40,750 & 72 & 8 & 4 \\
    & Common Sense & 21,857 & 38 & 4 & 2 \\
    & \textit{Subtotal} & \textit{384,136} & \textit{713} & \textit{77} & \textit{38} \\
    \hline
    \multirow{8}{0.07\textwidth}{\textbf{Computer Science} \cite{bruch2024_vector_retrieval}\cite{yu2022_common_information}\cite{orabona2019_online_learning}\cite{simeone2017_ml_intro}\cite{abbe2017_sbm_survey}\cite{bubeck2014_convex_algorithms}\cite{bussemaker2025_sao_strategies}}
    & Modern Introduction to Online Learning & 103,145 & 171 & 20 & 10 \\
    & Common Information, Noise Stability & 79,782 & 133 & 15 & 7 \\
    & Foundations of Vector Retrieval & 65,275 & 100 & 12 & 6 \\
    & Machine Learning Fundamentals & 60,133 & 101 & 11 & 5 \\
    & Community Detection Algorithms & 42,630 & 71 & 8 & 4 \\
    & Convex Optimization & 33,051 & 55 & 6 & 3 \\
    & System Architecture & 18,039 & 33 & 0 & 0 \\
    & \textit{Subtotal} & \textit{402,055} & \textit{664} & \textit{72} & \textit{35} \\
    \hline
    \multirow{2}{0.07\textwidth}{\textbf{Science}   \cite{openstax_microbiology}}
    & Microbiology Textbook & 397,994 & 678 & 80 & 20 \\
    & \textit{Subtotal} & \textit{397,994} & \textit{678} & \textit{80} & \textit{20} \\
    \hline
    \multicolumn{2}{|l|}{\textbf{Total Dataset}} & \textbf{1,581,333} & \textbf{2,734} & \textbf{309} & \textbf{133} \\
    \hline
    \end{tabular}
    }
\end{table}

We systematically generated 3,176 question-answer (QA) pairs across texts. The questions are open-ended to simulate how a student or instructor might ask about the material. Crucially, we categorized each question into one of three types based on the scope of information required to answer:

\begin{itemize}
\item \textbf{Specific Questions}: These are narrow questions answerable by a single paragraph ($\approx$ 500 words). These typically focus on a specific fact or definition. In the \textit{Microbiology} textbook\cite{openstax_microbiology}, a specific question is: ``Which genera of soil bacteria are involved in the denitrification process?"

\item \textbf{Sectional Questions}: These questions require aggregating information from multiple paragraphs (i.e. a chapter). For example, ``How did U.S. President Wilson's approach to negotiation affect the Paris Peace Conference?" Answering this requires pulling points spread across a section of \textit{Economic Consequences of Peace}\cite{keynes_economic_consequences}.  

\item \textbf{Thematic Questions}: These are broad questions that relate to overarching themes or cross-cutting concepts. For instance, a question from \textit{Moby Dick}\cite{melville_mobydick}: ``What does the Whaleman’s Chapel represent?" The system must draw on understanding from the novel as a whole, involving reasoning over tens of thousands of words.
\end{itemize}

The question generation pipeline handles long texts by breaking them into manageable pieces and iteratively summarizing and filtering content. The pipeline can be seen in Fig.~\ref{fig:QAGen_flowchart}, and operates as follows:

\begin{figure*}[t]
\centering
\includegraphics[width=0.9\textwidth]{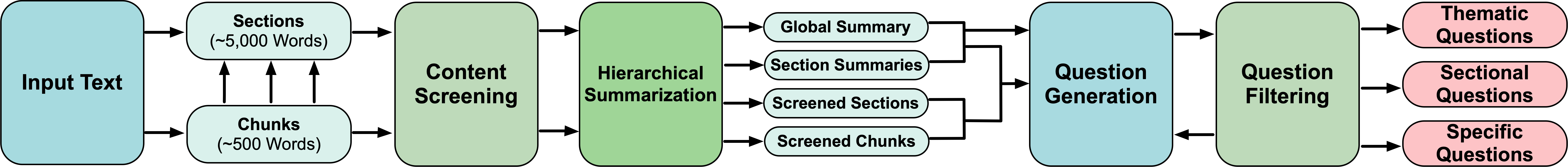}
\caption{EduScopeQA Dataset Question Generation Pipeline}
\label{fig:QAGen_flowchart}
\end{figure*}

\begin{enumerate}
    \item \textbf{Chunking/Sectioning}: Each text is split into chunks. Sets of ten consecutive chunks are grouped into sections. 
    \item \textbf{Content Screening}: GPT-4.1 examines each chunk and section, filtering those that consist of irrelevant or extraneous content such as front matter, textbook objectives, publication info, and acknowledgments.
    \item \textbf{Hierarchical Summarization}: For each section, GPT-4.1 generates a concise summary capturing its key points. If the combined content still exceeds about 35,000 characters, the summaries are sectioned and summarized again. This recursive summarization continues until we obtain a global summary of the entire text. This hierarchy (chunk → section summary → global summary) compresses the text while preserving important information at multiple levels.
    \item \textbf{Specific/Sectional Question Generation}: From the screened sections and chunks, a subset is randomly sampled. For each chosen one, GPT-4.1 is provided a summary of the surrounding sections and the global summary. With this, GPT-4.1 generates QA pairs specific to the current section while understanding the broader context. This helps produce coherent questions that remain answerable from the target information. 
    \item \textbf{Thematic Question Generation}: Thematic questions are produced using the global summary, prompting GPT-4.1 to cover the main themes or topics of the entire text.
    \item \textbf{Filtering and Review}: Each QA pair is passed through GPT-4.1 again filtering out unanswerable or trivial questions. Steps 4/5 are repeated accordingly.
\end{enumerate}

These dataset construction decisions were taken to mimic educational applications. Notably, no QA dataset exists that varies subject and question scope in this way. By evaluating performance across these categories, we can analyze whether a method excels at pinpointing facts but struggles with synthesis, or vice versa. This granularity is valuable for educational applications: a simple AI tutor might be fine if it can recall facts, but true learning assistance requires the system to also explain broad themes correctly.

\subsection{Experiment}
The full corpus of text for each subject was uploaded (for OpenAI RAG) and indexed (for GraphRAG Local and GraphRAG Global) using GPT-4.1-Mini. Then, the questions were used as input to querying both systems one at a time, again using GPT-4.1-Mini. The answers of each system were captured and then evaluated for a comparison analysis below.

\subsection{Evaluation}
We evaluated each answer using an ``LLM-as-a-Judge" technique, which has been shown to achieve human-level consistency for open-ended answers \cite{llmeval_intro}. For each question, we compared every pair of systems. For each comparison, we used a fixed prompt template that presented GPT-4.1-Nano with the two candidate answers (‘Option A’ and ‘Option B’) and an instruction to choose the better response or declare a tie at \textit{a given criteria}. Four complementary criteria were chosen:
\begin{enumerate}
    \item \textbf{Comprehensiveness}: Does the answer cover all relevant points and facets of the question? 
    \item \textbf{Directness}: Is the answer succinct, and to the point without unnecessary digression? 
    \item \textbf{Faithfulness}: Is the answer faithful to the ground truth? 
    \item \textbf{Learnability}: How well does the answer help a student learn or understand the topic? This criterion covers clarity of explanation, quality of reasoning, and pedagogical value.
\end{enumerate}

These criteria build on recent RAG QA studies, such as \cite{lightrag} and Microsoft GraphRAG's introduction paper \cite{microsoft}, that use LLM-as-a-Judge for their evaluations. The criteria were adapted to assess not only factual accuracy but also instructional clarity and pedagogical value. By including Directness and Comprehensiveness, the rubric inherently counterbalances verbosity and coverage biases, enabling fairer, more meaningful comparisons.

Empirically, LLM-as-a-judge evaluations are prone to positional bias, a systematic tendency of LLMs to prefer an answer due to its position in the prompt\cite{llm_evaluator}. To mitigate this, for each pairwise comparison, we followed an AB-BA strategy, passing in the two candidate answers as ``Option A" and ``Option B", and then swapping them, feeding the original ``Option B" as A and ``Option A" as B. We then aggregated the two judgments into a final winner or ‘tie’. Studies have shown that such a strategy balances out the positional skew\cite{humans_or_llm}.

\subsection{Discussion}

In order to comparatively analyze the LLM-as-a-judge evaluation, we used ``Win Rate" as follows

\begin{equation}
    W_A = \frac{w_A + 0.5 \cdot t_A}{n}
    \label{eq:win_rate}
    \end{equation}

where $W_A$ is the win rate for AI system $A$, $w_A$ is the number of wins for system $A$, $t_A$ is the number of ties for system $A$, and $n$ is the total number of pairwise comparisons being considered, leading to a score out of 1, where higher values mean a higher win percentage at a certain criterion.

\begin{figure*}[t]
\centering
\includegraphics[width=0.32\textwidth]{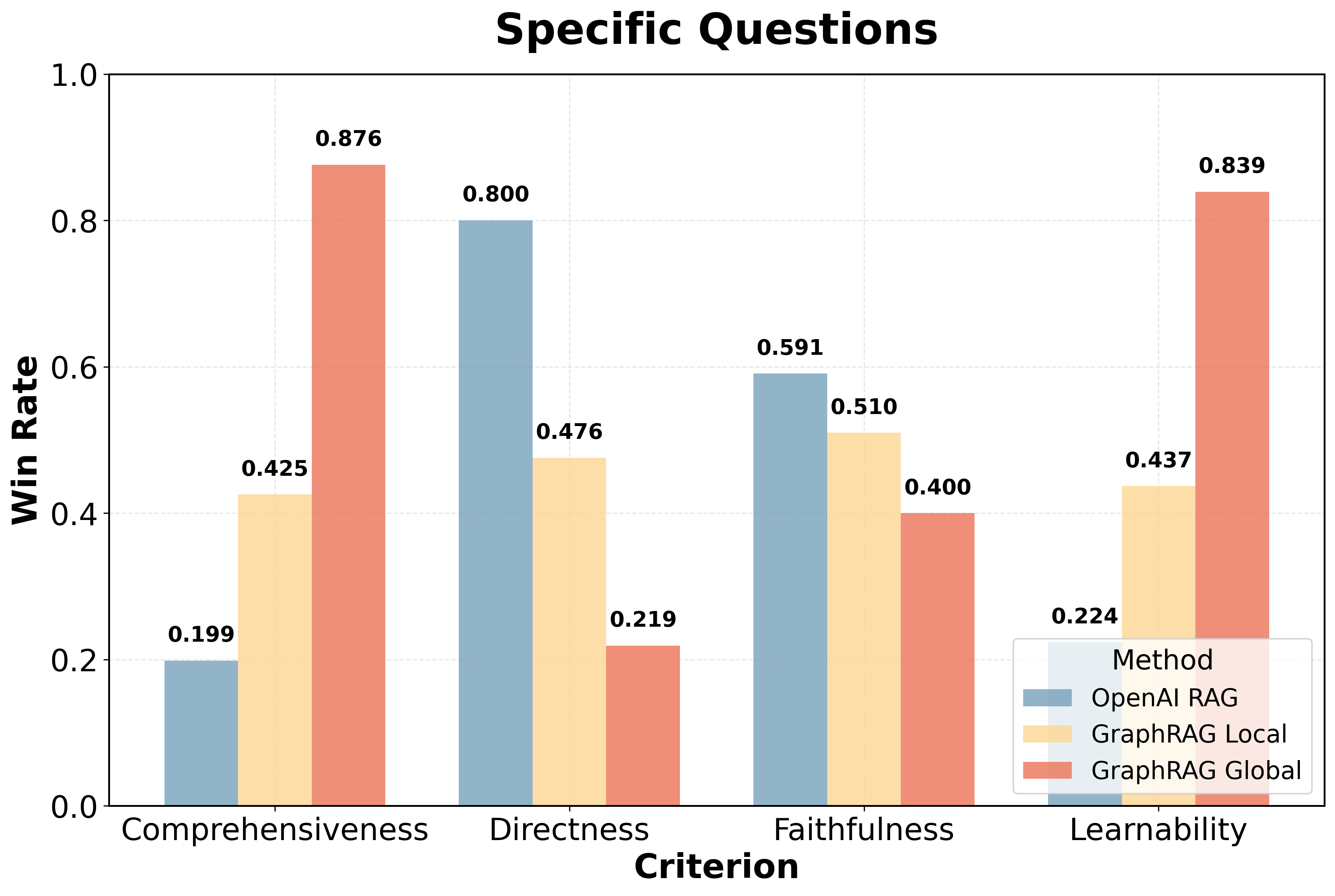}
\hfill
\includegraphics[width=0.32\textwidth]{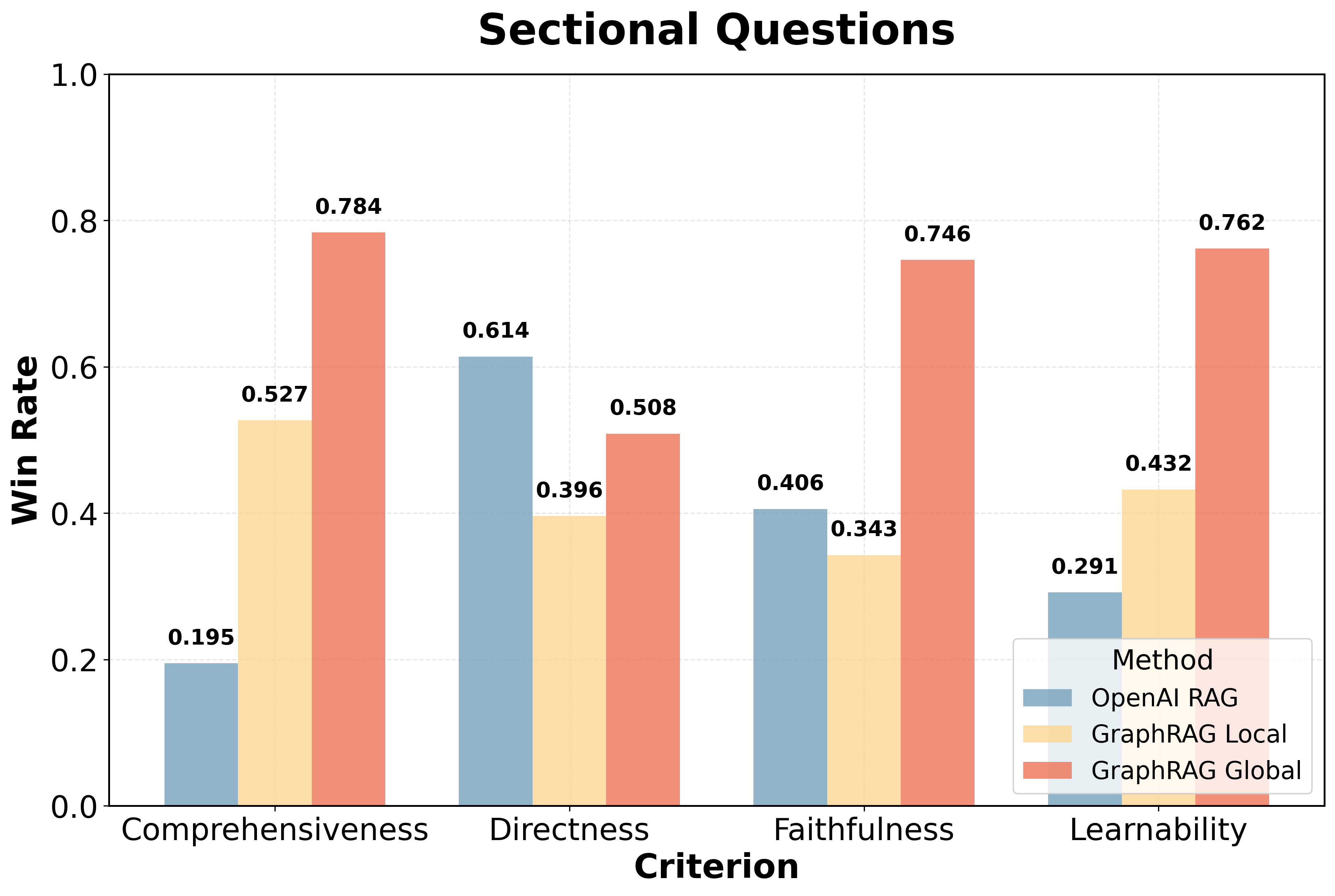}
\hfill
\includegraphics[width=0.32\textwidth]{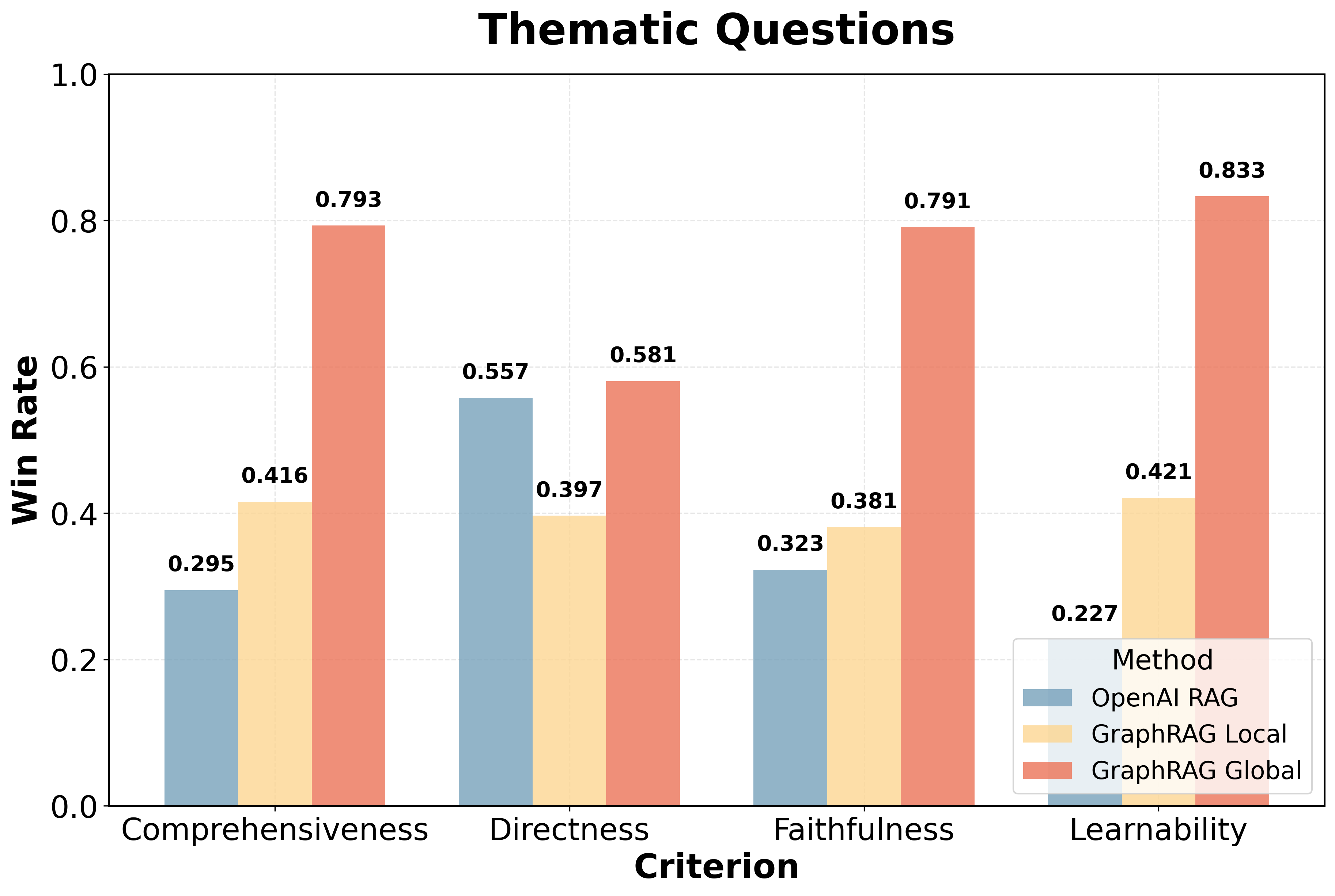}
\caption{Case Study 1 Results: Average Win Rates across question types and criteria}
\label{fig:question_types}
\end{figure*}

The full results of our experiment can be seen in Table~\ref{tab:all_methods_results}, and a visual comparison of win rate across question types can be seen in Fig.~\ref{fig:question_types}.

\begin{table*}[!htb]
    \centering
    \caption{Average Win Rates - All Methods}
    \label{tab:all_methods_results}
    \resizebox{\textwidth}{!}{%
    \begin{tabular}{l|c|ccc|ccc|ccc|ccc}
    \toprule
    \textbf{Metric} & \textbf{Method} & \multicolumn{3}{c|}{\textbf{Computer Science}} & \multicolumn{3}{c|}{\textbf{History}} & \multicolumn{3}{c|}{\textbf{Literature}} & \multicolumn{3}{c}{\textbf{Science}} \\
    \addlinespace[0.5em]
     & & \textbf{Specific} & \textbf{Sectional} & \textbf{Thematic} & \textbf{Specific} & \textbf{Sectional} & \textbf{Thematic} & \textbf{Specific} & \textbf{Sectional} & \textbf{Thematic} & \textbf{Specific} & \textbf{Sectional} & \textbf{Thematic} \\
    \midrule
    \multirow{3}{*}{\textbf{Comprehensiveness}} & \textbf{OpenAI RAG} & 0.200 & 0.264 & 0.429 & 0.214 & 0.221 & 0.237 & 0.204 & 0.138 & 0.263 & 0.176 & 0.156 & 0.250 \\
     & \textbf{GraphRAG Local} & 0.422 & 0.451 & 0.243 & 0.429 & 0.506 & 0.382 & 0.407 & 0.600 & 0.463 & 0.444 & 0.550 & 0.575 \\
     & \textbf{GraphRAG Global} & \underline{0.879} & \underline{0.812} & \underline{0.829} & \underline{0.857} & \underline{0.766} & \underline{0.868} & \underline{0.889} & \underline{0.750} & \underline{0.825} & \underline{0.880} & \underline{0.806} & \underline{0.650} \\
    \cmidrule{2-14}
    \multirow{3}{*}{\textbf{Directness}} & \textbf{OpenAI RAG} & \underline{0.780} & 0.556 & 0.312 & 0.875 & 0.506 & 0.618 & \underline{0.870} & \underline{0.600} & \underline{0.700} & \underline{0.676} & \underline{0.794} & 0.600 \\
     & \textbf{GraphRAG Local} & 0.421 & 0.444 & 0.562 & 0.482 & 0.396 & 0.224 & 0.426 & 0.594 & 0.425 & 0.574 & 0.150 & 0.375 \\
     & \textbf{GraphRAG Global} & 0.298 & 0.556 & \underline{0.625} & 0.143 & \underline{0.597} & \underline{0.684} & 0.204 & 0.312 & 0.388 & 0.231 & 0.569 & \underline{0.625} \\
    \cmidrule{2-14}
    \multirow{3}{*}{\textbf{Accuracy}} & \textbf{OpenAI RAG} & \underline{0.599} & 0.562 & 0.471 & 0.625 & 0.442 & 0.395 & \underline{0.815} & 0.494 & 0.300 & 0.324 & 0.125 & 0.125 \\
     & \textbf{GraphRAG Local} & 0.452 & 0.208 & 0.214 & \underline{0.661} & 0.188 & 0.197 & 0.435 & 0.394 & 0.438 & 0.491 & 0.581 & 0.675 \\
     & \textbf{GraphRAG Global} & 0.451 & \underline{0.764} & \underline{0.871} & 0.214 & \underline{0.851} & \underline{0.868} & 0.250 & \underline{0.588} & \underline{0.750} & \underline{0.685} & \underline{0.781} & 0.675 \\
    \cmidrule{2-14}
    \multirow{3}{*}{\textbf{Learnability}} & \textbf{OpenAI RAG} & 0.209 & 0.319 & 0.175 & 0.232 & 0.253 & 0.171 & 0.278 & 0.163 & 0.338 & 0.176 & 0.431 & 0.225 \\
     & \textbf{GraphRAG Local} & 0.451 & 0.389 & 0.450 & 0.464 & 0.240 & 0.447 & 0.398 & 0.588 & 0.362 & 0.435 & 0.512 & 0.425 \\
     & \textbf{GraphRAG Global} & \underline{0.840} & \underline{0.771} & \underline{0.825} & \underline{0.804} & \underline{0.994} & \underline{0.882} & \underline{0.824} & \underline{0.725} & \underline{0.775} & \underline{0.889} & \underline{0.556} & \underline{0.850} \\
    \bottomrule
    \end{tabular}
    }
    \end{table*}

\textbf{GraphRAG Global dominates Broad Queries and Pedagogical Criteria}. GraphRAG Global achieves the highest win rates in Faithfulness in Sectional and Thematic questions. Its multi-hop retrieval, leveraging the hierarchical structure of knowledge graphs, synthesizes dispersed information effectively. GraphRAG Global also dramatically scores well in Comprehensiveness and Learnability across question types and subjects. This shows that its long-range connectivity allows the LLM to generate comprehensive, pedagogically rich responses that support learning. This supports other results that global graph-based RAG excels at ``broad" questions and implies its effectiveness as a tool for teaching concepts.

\textbf{OpenAI RAG excels at Specific Queries and Directness}. While GraphRAG Global scores well on the pedagogical criteria for specific questions, it is overshadowed by OpenAI RAG in the other criteria. Since specific factual queries often reside in a single retrieved snippet, GraphRAG Global struggles to capture these details with high-level summarization. Consequently, OpenAI RAG is optimal for ``flashcard" applications, quick glossary references, and scenarios requiring immediate, precise answers.

\textbf{GraphRAG Local acts as a Competent Bridge}. And in between, GraphRAG Local scores better on Faithfulness and Directness at specific questions than GraphRAG Global, but performs better at pedagogical criteria (Comprehensiveness and Learnability) than OpenAI RAG across the board. By constraining graph reasoning to local neighborhoods, it yields more complete responses when answers span several paragraphs, but leads to limitations in broad thematic coverage in comparison with GraphRAG Global.

\textbf{Subject Variations}. While overall patterns hold, there are nuances between subjects. For example, in Computer Science, the gap between OpenAI RAG and GraphRAG Global for Faithfulness in sectional (0.562 vs. 0.764) and thematic (0.625 vs. 0.812) is much smaller compared with Literature (0.125 vs. 0.781 and 0.125 vs. 0.675 respectively). This underscores that for fictional novels, which make up the Literature corpus, truly global cues (motifs, narrative arcs) are widely dispersed and benefit greatly from global retrieval, whereas for technical papers, which make up the Computer Science corpus, factual claims make up the bulk of subject matter and impose fewer integrative demands. Science follows results similar to Computer Science, while History falls nearer to Literature.

\section{Case Study 2 (CS2)}

For Case Study 2, we used the \textbf{KnowShiftQA} (KSQA) dataset, which is specifically designed to test a retrieval system’s ability to prioritize provided source material over an LLM’s internal knowledge\cite{knowshift}. The dataset comprises five textbooks from secondary education level subjects: Physics, Chemistry, Biology, Geography, and History. In each textbook, certain pieces of factual information have been systematically altered to simulate hypothetical knowledge updates. Importantly, these changes are done in a coherent way so that the surrounding context in the textbook remains plausible and internally consistent. These alterations include changing the textbook to claim that ``Night-vision goggles detect ultraviolet light," from infrared light (in reality). So, a question like ``What type of light is detected by night-vision goggles?" would test a QA system’s faithfulness to the input corpus much better. This is an important control for testing retrieval systems, as it renders some of the LLM’s latent parametric knowledge incorrect, forcing it to rely on the retrieval. 

The dataset contains 3,005 QA pairs. Each question is a multiple-choice question (MCQ) with one correct answer (which corresponds to the altered fact) and several distractors (including the real-world fact).

\subsection{Experiment}

We also compare how document length matters since course material range from short articles and handouts to textbooks and novels. We boost the KSQA dataset by creating three experimental settings (KSQA provides alignment of questions to the source text): 

\begin{itemize}
    \item \textbf{Short-Retrieval}: Only the chunk from which each question was generated was provided as the input corpus ($\approx$ 315 words).
    \item \textbf{Medium-Retrieval}: Only the 30 chunks surrounding the one each question was generated from were provided as the input corpus ($\approx$ 9.5 K words).
    \item \textbf{Full-Retrieval}: The entire textbook was provided as the input corpus.
\end{itemize}

In all three settings, the appropriate text was uploaded (for OpenAI RAG) and indexed (for GraphRAG Local and GraphRAG Global) using GPT-4.1-Mini. Then, both systems were queried one at a time, again using GPT-4.1-Mini. The MCQ format was handled by including the options and prompting the model to choose an answer.

\begin{figure}[!htb]
\centering
\includegraphics[width=\columnwidth]{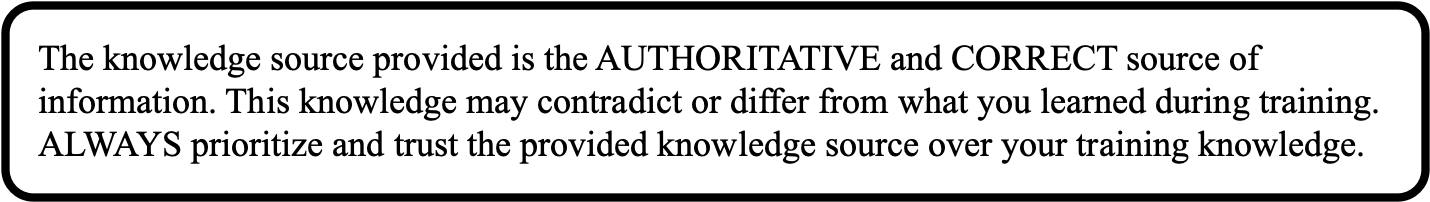}
\caption{LLM Prompt for Case Study 2}
\label{fig:ksqa_prompt}
\end{figure}

It is important to note that a prompt instructing the LLM to ignore factuality of answers was passed in at query-time to all three systems. The full prompt is seen in Fig.~\ref{fig:ksqa_prompt}. Furthermore, GraphRAG automatically generates prompts that are used to query the knowledge graph at query-time. Some of these graphs contain wording encouraging outside reasoning, such as ``The response may also include relevant real-world knowledge outside the dataset". To standardize the experiment, these sentences were removed and replaced with the same prompt.

\begin{figure*}[!t]
    \centering
    \includegraphics[width=0.32\textwidth]{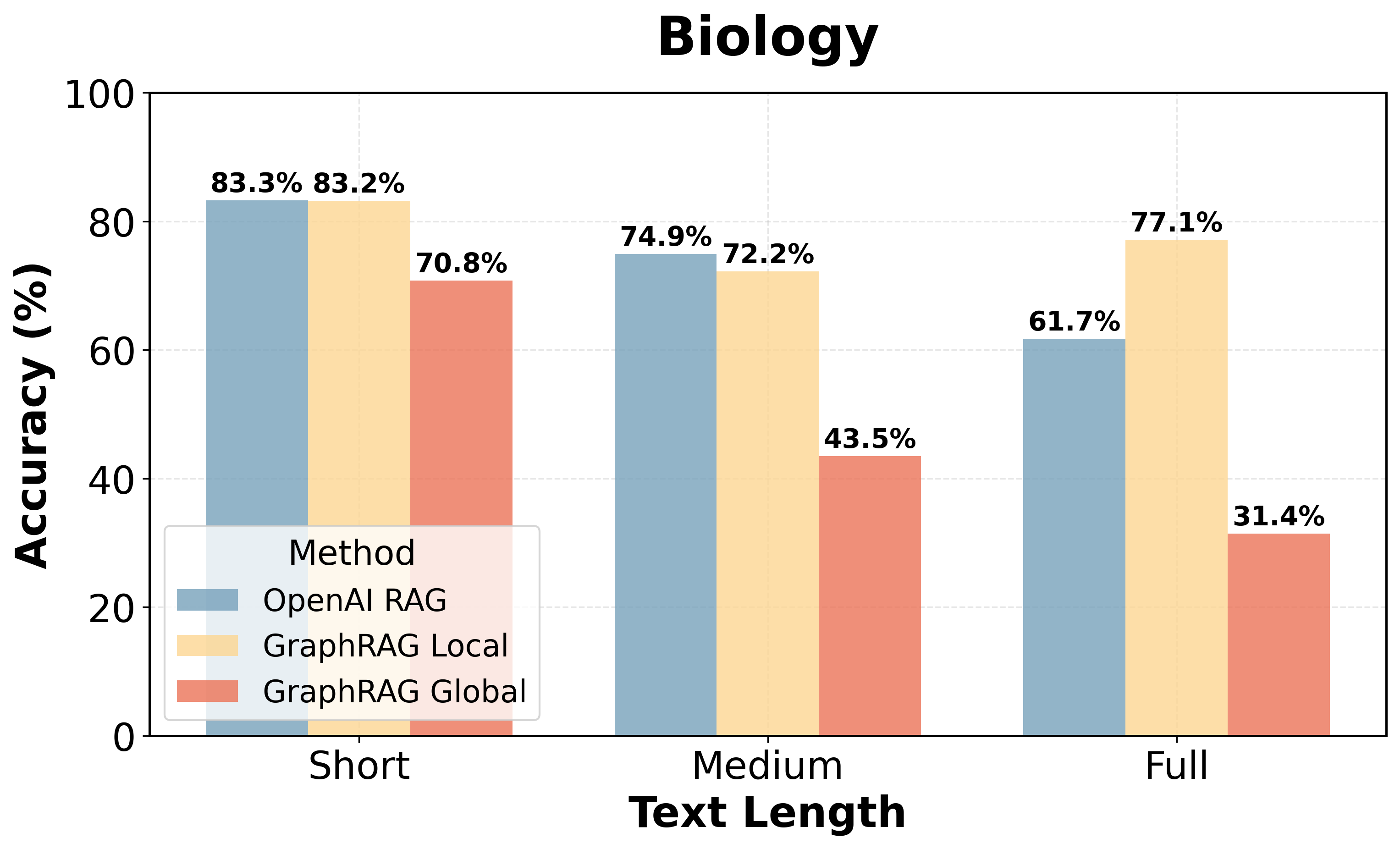}
    \hfill
    \includegraphics[width=0.32\textwidth]{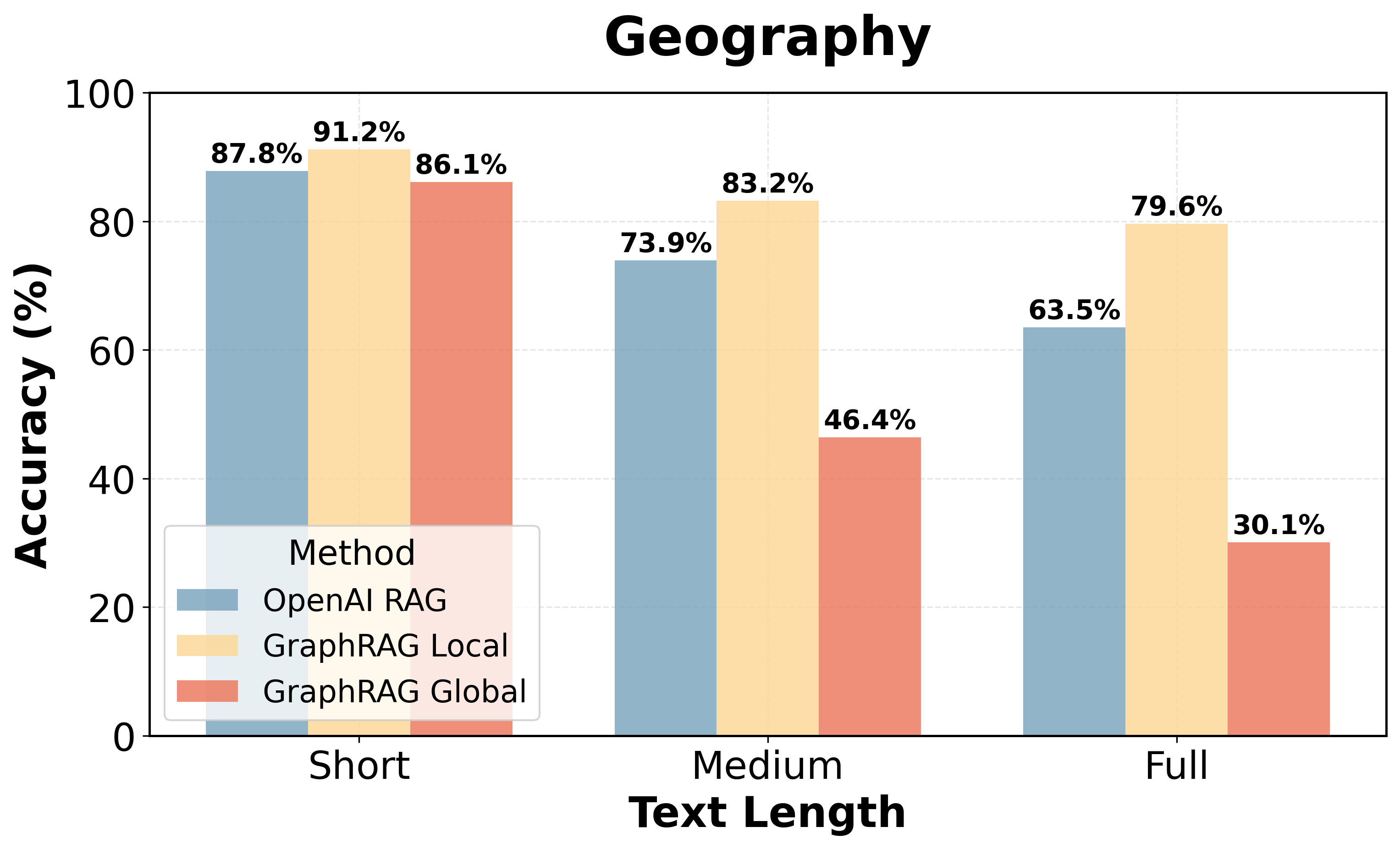}
    \hfill
    \includegraphics[width=0.32\textwidth]{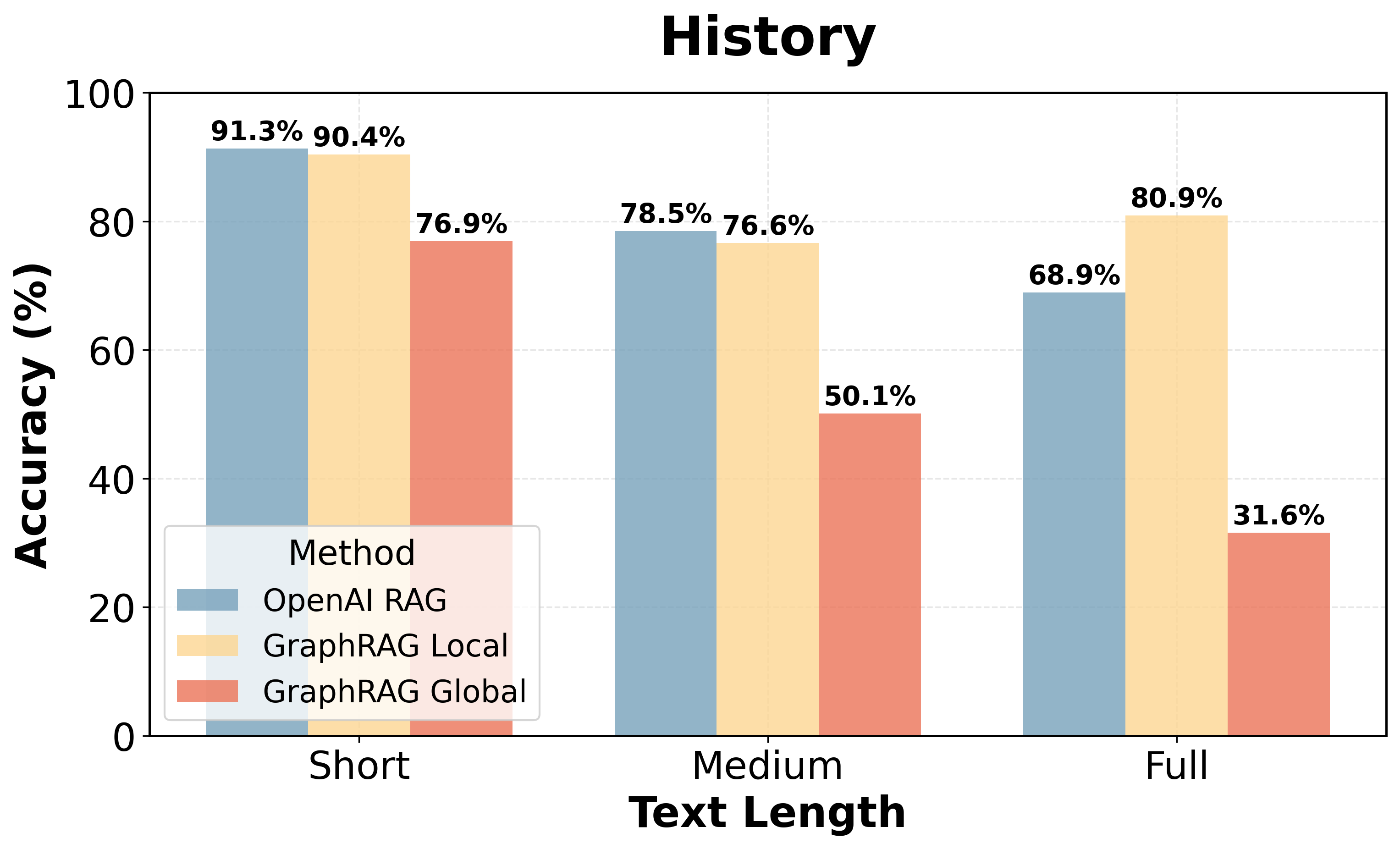}
    \hfill
    \includegraphics[width=0.32\textwidth]{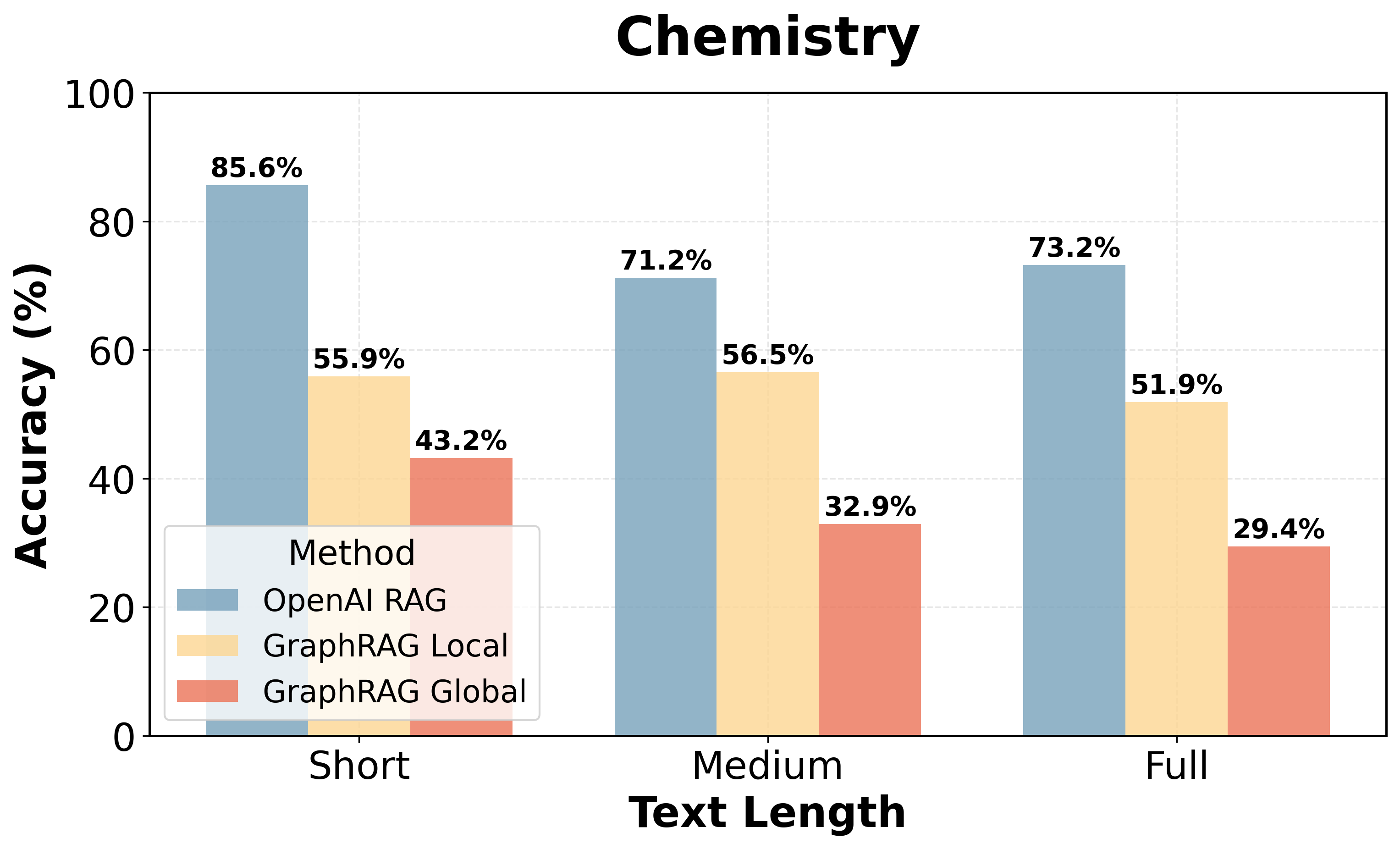}
    \includegraphics[width=0.32\textwidth]{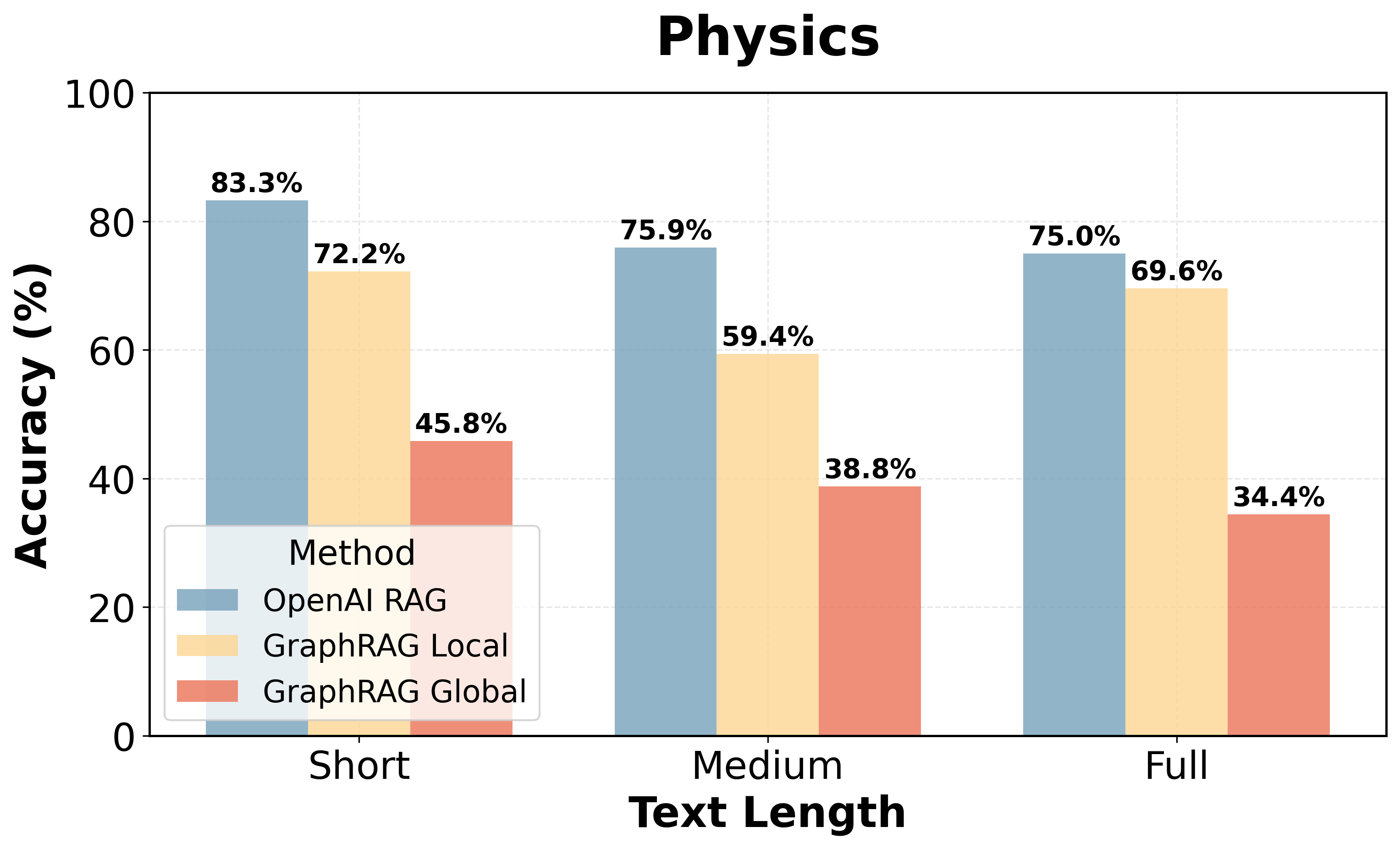}
    \caption{Case Study 2 Results: Average Accuracy by Subject and Retrieval Scope. \textbf{Large textbooks} - Biology (258K words), History (146K), Geography (165K); \textbf{Smaller Textbooks} - Chemistry (77K), Physics (68K)}
    \label{fig:ksqa_results}
\end{figure*}

\subsection{Evaluation}
We checked if the chosen answer was the correct (altered) one by performing a fuzzy matching using a GPT-4.1-Nano evaluation. We measured accuracy in terms of percentage of questions answered correctly.

\subsection{Discussion}
Our findings (Fig.~\ref{fig:ksqa_results}) reveal clear performance patterns influenced by corpus size and retrieval scope:

\textbf{GraphRAG Local for Large, Dense Corpora}. In full-retrieval, especially in the larger Biology (258 K words), History (146 K words), and Geography (165 K words) textbooks, GraphRAG Local consistently outperforms both OpenAI RAG and GraphRAG Global. Its local graph structure efficiently identifies precise factual information amidst large volumes of potentially distracting content. Previous works, and even Case Study 1 imply that vector-based RAG performs better on specific, factual questions. We show that for a large corpus of dense facts, GraphRAG Local is better at identifying minutiae and maintaining strict adherence to curriculum content.

\textbf{OpenAI RAG in Smaller Corpora}. In smaller texts such as Chemistry (77 K words) and Physics (68 K words), OpenAI RAG closely matches or slightly outperforms GraphRAG Local. Across the medium and short retrieval conditions, all systems converge on high scores, but OpenAI RAG generally leads. With reduced corpus size, vector retrieval precision effectively compensates for lack of structured multi-hop capabilities, indicating diminishing returns for graph construction (which can actually add distractors).

\section{Overall Discussion}
For both case studies, we recorded the number of LLM calls during indexing, the total indexing time, and the query time (Table~\ref{tab:indexing_costs}). \begin{table}[b]
    \centering
    \caption{Avg Indexing Per Subject and Query Costs by Case Study (CS)}
    \label{tab:indexing_costs}
    \resizebox{0.8\columnwidth}{!}{
    \begin{tabular}{l|c|cc}
    \toprule
    \textbf{Indexing} & \textbf{OpenAI RAG} & \multicolumn{2}{c}{\textbf{GraphRAG}} \\
    & \textbf{Time (s)} & \textbf{Time (s)} & \textbf{LLM Calls} \\
    \midrule
    CS1 & 11.43 & 2,142.22 & 4,025.25 \\
    CS2-Full & 5.87 & 1,078.12 & 2,038.4 \\
    CS2-Medium & 3.82 & 186.61 & 112.45 \\
    CS2-Short & 3.53 & 48.12 & 10.28 \\
    \bottomrule
    \end{tabular}
    }

    \vspace{1mm}

    \resizebox{0.95\columnwidth}{!}{%
    \begin{tabular}{l|c|c|c}
    \toprule
    \textbf{Querying} & \textbf{OpenAI RAG} & \textbf{GraphRAG Local} & \textbf{GraphRAG Global} \\
    & \textbf{Time (s)} & \textbf{Time (s)} & \textbf{Time (s)} \\
    \midrule
    CS1 & 4.71 & 36.50 & 70.12 \\
    CS2-Full & 5.00 & 35.57 & 39.41 \\
    CS2-Medium & 3.92 & 34.61 & 42.12 \\
    CS2-Short & 3.07 & 9.43 & 9.95 \\
    \bottomrule
    \end{tabular}
    }
\end{table}. We use \textit{LLM calls as a proxy for monetary cost} for generalization and normalization since billing depends on model provider and model variant pricing.

Notably, GraphRAG (Local and Global were indexed together) required substantial computational resources during indexing for entity and relationship extraction, whereas OpenAI RAG was lightning fast with zero additional LLM calls, as embedding is handled internally by OpenAI's infrastructure. The indexing burden for GraphRAG also scaled with document length (short, medium, full), whereas OpenAI RAG remained relatively stable. GraphRAG's one-time indexing of an average textbook from Case Study 1 required 35 minutes and 4,000 LLM calls. This overhead can exceed schools' budget if done regularly, particularly with costlier LLMs. This was even more applicable while querying, where GraphRAG Global (and Local to a lesser extent) scaled significantly.

Altogether, this provides a more complete discussion of pedagogical deployment. \textbf{(1) Vector-based RAG is excellent for most cases}, especially when students need individual, rapid, pinpoint responses such as targeted explanations about a paragraph or quick glossary lookups. Its low latency and ease of setup make it ideal for embedding into general chatbots without overburdening school IT resources. \textbf{(2) GraphRAG’s high initial costs can be justified when a particular corpus can be indexed and shared across users over time}. When the goal is to support essay prompts or seminar discussions centered on a classroom text, where students benefit from rich, concept-spanning explanations, \textbf{(3) GraphRAG Global provides the most coherent, curriculum-aligned narratives}. Despite higher runtime costs, this method can be justified in settings where depth of understanding matters most. For large, evolving textbooks, question banks, or multiple-choice questions, \textbf{(4) GraphRAG Local offers accuracy and context-sensitivity}. Its tight adherence to the provided material is critical when ensuring that answers align exactly with the latest curriculum standards or exam specifications. 

\subsection{Proof of Concept Branching System}

\begin{figure}[t]
\centering
\includegraphics[width=0.85\columnwidth]{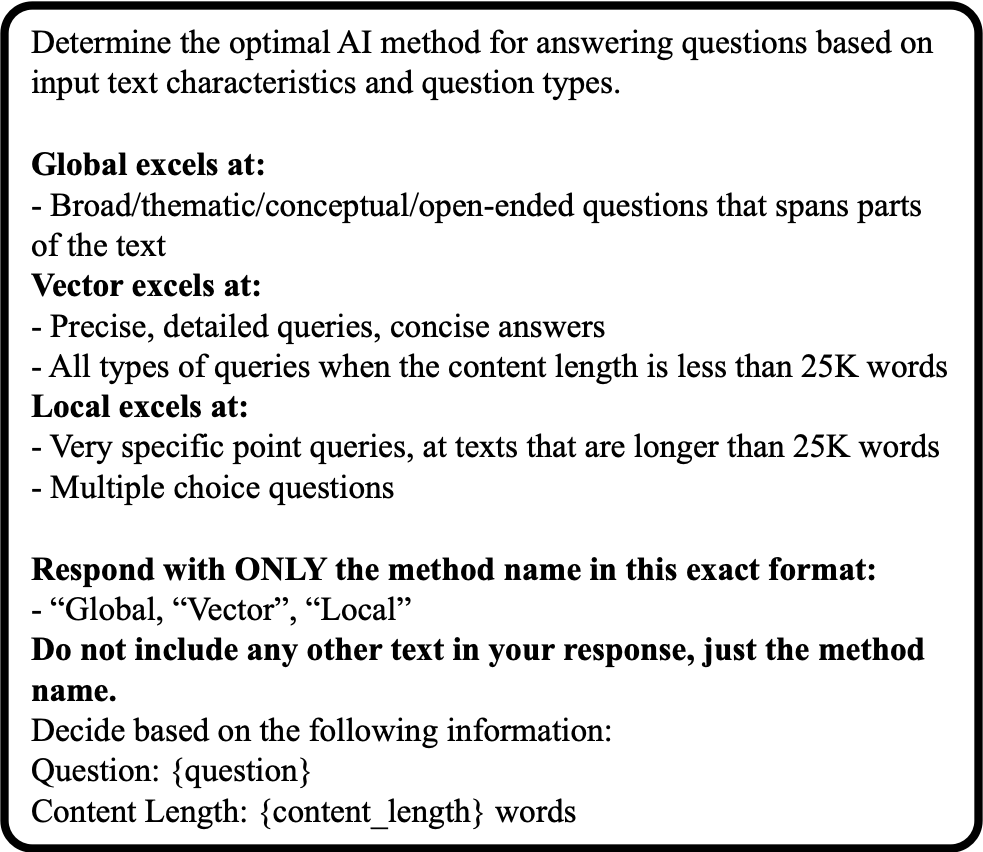}
\caption{Branching prompt to choose the optimal retrieval method}
\label{fig:brancher_prompt}
\end{figure}

Although separating the deployments of the systems by use case is an option, we show as a proof of concept a lightweight branching system which routes incoming queries based on complexity, scope, and corpus size to the appropriate retrieval system. We use an initial GPT-4.1-Nano call with the prompt shown in Fig.~\ref{fig:brancher_prompt}. The LLM is instructed to choose based on a short description of each system’s strengths.

\begin{figure}[b]
\centering
\includegraphics[width=0.65\columnwidth]{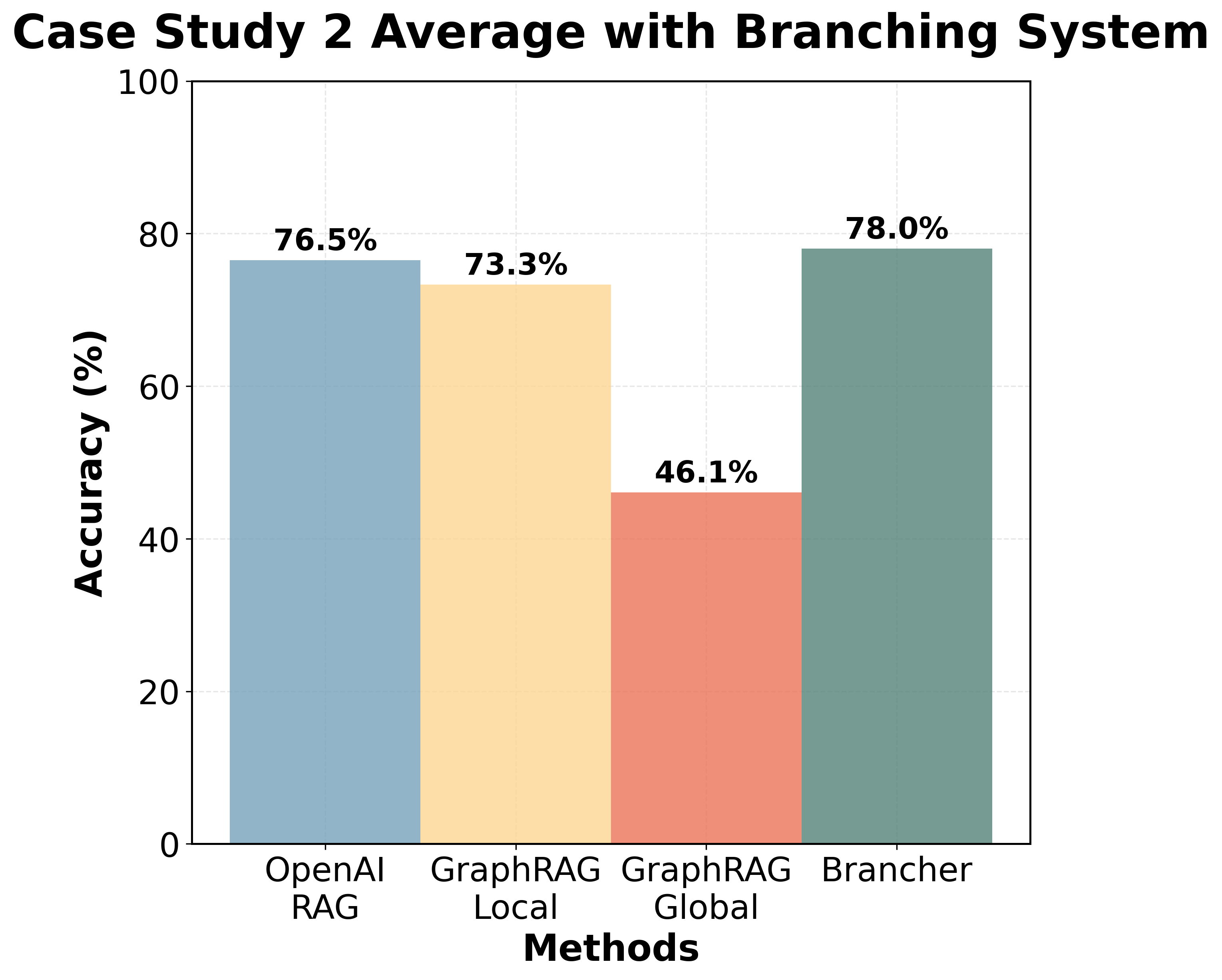}
\caption{Case Study 2 Results with Branching System, averaged across subjects and retrieval scopes}
\label{fig:brancher_accuracy}
\end{figure}

As a basic test, we repeated Case Study 1, but with the branching system as an addition. Each question and its input text were first passed through the LLM prompt in Fig.~\ref{fig:brancher_prompt}, and then the chosen system’s previous response was picked. We show the averaged results of the LLM-as-judge evaluation, \textit{without} converting to win-rates, in Table~\ref{tab:brancher_custom_results}. We also repeated Case Study 2 with the branching system, where the options of the MCQ were passed along with the question. Overall results, averaged across subjects and retrieval scopes, can be seen in Fig.~\ref{fig:brancher_accuracy}.
\begin{table}[b]
  \centering
  \caption{Case Study 1 Results with Branching System:\\ Averaged LLM-as-a-judge Win Percentages, \\ Branching Systems vs. the rest}
  \label{tab:brancher_custom_results}
  \resizebox{0.8\columnwidth}{!}{
    \begin{tabular}{l|c|c|c}
      \toprule
      \textbf{Metric}
      & \makecell{\textbf{OpenAI}\\\textbf{RAG}}
      & \makecell{\textbf{GraphRAG}\\\textbf{Local}}
      & \makecell{\textbf{GraphRAG}\\\textbf{Global}} \\
      \midrule
      \textbf{Comprehensiveness} & \underline{72.4\%} & \underline{67.6\%} & 37.0\% \\
      \textbf{Directness}        & 39.2\% & \underline{84.0\%} & \underline{66.1\%} \\
      \textbf{Faithfulness}          & \underline{68.5\%} & \underline{79.8\%} & \underline{60.2\%} \\
      \textbf{Learnability}      & \underline{80.1\%} & \underline{74.3\%} & 33.4\% \\
      \bottomrule
    \end{tabular}
  }
\end{table}
In Case Study 1, the branching system achieved the highest overall faithfulness scores of any single system, reflecting its ability to invoke OpenAI RAG for specific queries and GraphRAG Global for broader questions. By conditionally choosing a system according to question granularity, the brancher harnesses the strengths of both approaches. On criteria such as learnability and directness, however, the branching system occupies an intermediate position: it outperforms standalone OpenAI RAG or GraphRAG respectively in avoiding extreme weaknesses, but does not match each system’s highest strengths, GraphRAG Global’s rich narratives or OpenAI RAG’s most concise responses. This underscores a natural trade-off when balancing across question types.
For Case Study 2, it is evident that the branching system was able to take advantage of OpenAI RAG's strengths in shorter retrieval scopes and GraphRAG Local's higher accuracy in the larger corpora. Although the net gain in multiple-choice accuracy was more modest than in our open-ended evaluation, it displays that a branching system that dynamically chooses the best system situationally can be a useful tool to improve the performance of a QA system.

The resource costs of the branching system in both case studies are shown in Table~\ref{tab:brancher_costs}. Although costs are much lower than a pure GraphRAG system, they remain higher than a pure OpenAI RAG system, suggesting that further optimization is needed to make the branching system more cost-effective. Importantly, our cost calculations conservatively include indexing overhead for every routed query; in practice, schools can amortize GraphRAG’s setup costs by persisting indexed corpora across class cohorts and indexing during off-hours. Such optimizations would further narrow the price gap and enable sustainable, large-scale adoption of graph-based or dynamic RAG strategies in educational settings.
\begin{table}[b]
    \centering
    \caption{Branching Indexing and Query Costs (CS - Case Study)}
    \label{tab:brancher_costs}
    \resizebox{0.6\columnwidth}{!}{
    \begin{tabular}{l|cc|c}
    \toprule
     & \multicolumn{2}{c}{\textbf{Indexing}} & \textbf{Querying} \\
    & \textbf{Time (s)} & \textbf{LLM Calls} & \textbf{Time (s)} \\
    \midrule
    CS1 & 1,378.11 & 2,582.04 & 44.94 \\
    CS2 & 360.01 & 676.07 & 14.11 \\
    \bottomrule
    \end{tabular}
    }
\end{table}
\section{Conclusion and Future Work}
This paper offers a comprehensive, education-focused comparison of vector and graph-based RAG methods across realistic classroom QA scenarios. After testing with EduScopeQA, a novel multi-subject, multi-scope open-ended dataset, and evaluating accuracy under knowledge shifts, we demonstrate clear pedagogical trade-offs for real-world deployment. 

OpenAI RAG excels at rapid, precise fact retrieval due to its low resource cost. But when the focus on a particular corpus of text allows for longer-term shared use, GraphRAG’s high initial costs can be justified with improved accuracy and pedagogical value. Global produces the richest, curriculum-aligned explanations for thematic inquiry and Local ensures the highest fidelity when adhering to dense, evolving textbook content or multiple-choice questions. 

Our proof-of-concept branching framework demonstrates that intelligently routing questions to the optimal retrieval paradigm can yield improved accuracy while avoiding each method's limitations and reducing unnecessary computational overhead. These insights guide future efforts to deploying RAG-augmented LLMs in instructional contexts: whether powering live study aids, guiding deep seminar discussions, or safeguarding the integrity of classroom teachings.

\textbf{Limitations and Future Work.} Next steps should center on classroom pilots and co-design studies with teachers and students to validate our evaluation’s alignment with actual educational outcomes. Second, our evaluation only accounted for textual class materials; this calls for pedagogical evaluations of multimodal and visual RAG pipelines for educational images and videos. Third, the branching mechanism itself can be made robust by accounting for more factors and performing more rigorous testing. With these future directions, we can close the gap between technical innovation and real-world classrooms, ensuring AI systems remain aligned with diverse curricula and pedagogical goals.

\bibliographystyle{IEEEtran}
\bibliography{refs}

\end{document}